\pdfoutput=1

\documentclass[11pt]{article}

\usepackage{acl}

\usepackage{times}
\usepackage{latexsym}
\usepackage{amsmath}
\usepackage{dsfont}
\usepackage{booktabs}
\usepackage{xcolor}
\usepackage{colortbl}

\usepackage[T1]{fontenc}

\usepackage[utf8]{inputenc}

\usepackage{microtype}
\usepackage{dirtytalk}
\usepackage{multirow}

\title{Greed is All You Need: An Evaluation of Tokenizer Inference Methods}

\author{Omri Uzan$^{\beta}$ \quad
  Craig W. Schmidt$^{\kappa}$ \quad
  Chris Tanner$^{\kappa\mu}$ \quad
  Yuval Pinter$^{\beta}$ \\ \\
  \begin{tabular}{cc}
      $^\beta$ Department of Computer Science & $^\kappa$ Kensho Technologies \\
      Ben-Gurion University of the Negev & $^\mu$ Massachusetts Institute of Technology \\
      Beer Sheva, Israel & Cambridge, MA, USA \\
  \texttt{\small \{omriuz@post, uvp@cs\}.bgu.ac.il} &
  \texttt{\small \{craig.schmidt, chris.tanner\}@kensho.com}
  \end{tabular}}

\begin{document}
\maketitle


\begin{abstract}
While subword tokenizers such as BPE and WordPiece are typically used to build vocabularies for NLP models, the method of decoding text into a sequence of tokens from these vocabularies is often left unspecified, or ill-suited to the method in which they were constructed.
We provide a controlled analysis of seven tokenizer inference methods across four different algorithms and three vocabulary sizes, performed on a novel intrinsic evaluation suite we curated for English, combining measures rooted in morphology, cognition, and information theory.
We show that for the most commonly used tokenizers, greedy inference performs surprisingly well; and that SaGe, a recently-introduced contextually-informed tokenizer, outperforms all others on morphological alignment.
\end{abstract}

\section{Introduction}
\label{sec:intro}
Modern NLP systems, including large language models (LLMs), typically involve an initial step of mapping raw input text into sequences of subword tokens. These tokens are selected from a large vocabulary of candidates that were produced from algorithms such as Byte-Pair Encoding~\cite[BPE;][]{sennrich-etal-2016-neural}, WordPiece~\cite{schuster2012japanese}, or UnigramLM~\cite{kudo-2018-subword}.

This process, which we refer to as the \textit{inference method} of tokenization, is critical as it determines how all text is represented and subsequently modeled.
Each inference method offers distinct mappings, and we assert that it is not well-understood how these methods differ in performance.
Furthermore, popular implementation packages such as Huggingface Tokenizers,\footnote{\scriptsize \url{https://huggingface.co/docs/tokenizers}} SentencePiece,\footnote{\scriptsize \url{https://pypi.org/project/sentencepiece}} and SubwordNMT\footnote{\scriptsize \url{https://github.com/rsennrich/subword-nmt}} often obfuscate or even restrict the choice of inference methods, making it unclear if inference-time decoding is compatible with the algorithm used to learn the tokenizer's vocabulary.
Moreover, it is yet to be determined whether such a match is ideal, or even necessary.

\begin{table}
    \centering
    \small
    \begin{tabular}{ll}
    \toprule
        Tokenizer$_{\text{inference mode}}$ & Segmentation \\
         \midrule
         BPE$_{\text{merges}}$ & \texttt{\textvisiblespace{}Ul tr am od ern}\\
         BPE$_{\text{longest prefix}}$ & \texttt{\textvisiblespace{}Ultra modern} \\
         \midrule
         UnigramLM$_{\text{likelihood}}$ & \texttt{\textvisiblespace{} U nprecedented}\\
         UnigramLM$_{\text{longest prefix}}$ & \texttt{\textvisiblespace{}Un precedent ed}\\
         \midrule
         SaGe$_{\text{longest prefix}}$ & \texttt{\textvisiblespace{}Inc once iva ble}\\
         SaGe$_{\text{likelihood}}$ & \texttt{\textvisiblespace{}In conceiv able} \\
         \bottomrule
    \end{tabular}
    \caption{Examples of words being segmented differently by various tokenizers (vocab size 32,000) using different inference modes on the same vocabulary. Each tokenizer's default mode is provided on top.}
    \label{tab:examples}
\end{table}

In \autoref{tab:examples} we present examples demonstrating how the prescribed inference methods of BPE, UnigramLM, and SaGe~\cite{yehezkel-pinter-2023-incorporating} do not necessarily provide the best segmentation for complex English words, even when good segments are available in the vocabulary.
BPE's out-of-the-box algorithm merges the cross-morphemic \texttt{am} sequence at an early stage, preventing the consideration of \texttt{ultra} and \texttt{modern} and condemning the downstream model to work with a representation learned for the first-person present form of `to be'.
UnigramLM's ablative algorithm enabled \texttt{nprecedented} (which crosses morpheme boundaries) to remain in its final vocabulary of tokens, while SaGe's greedy algorithm masks the boundaries of both the prefix \texttt{In} and the suffix \texttt{able}.
In all cases, an alternative inference method provides a more morphologically-aligned segmentation over the same vocabulary.

Previous work regarding subword tokenization mostly concerns developing vocabulary construction algorithms~\cite{sennrich-etal-2016-neural,schuster2012japanese,kudo-2018-subword,mielke2021between,yehezkel-pinter-2023-incorporating}, finding the optimal vocabulary size~\citep{gowda-may-2020-finding,gutierrez-vasques-etal-2021-characters}, building multilingual vocabularies~\citep{liang-etal-2023-xlm}, and using space positioning in the vocabulary tokens~\cite{gow-smith-etal-2022-improving,jacobs2022lost}.
Others analyze the effects of vocabularies, finding intricate relations between algorithm or vocabulary and downstream performance~\cite{bostrom-durrett-2020-byte,cognetta2024analysis}, information theory~\cite{zouhar-etal-2023-tokenization,cognetta-etal-2024-two-counterexamples}, cognitive plausibility~\cite{beinborn-pinter-2023-analyzing}, impact on society~\cite{ovalle2024tokenization}, or morphological alignment~\cite{klein-tsarfaty-2020-getting,hofmann-etal-2021-superbizarre,hofmann-etal-2022-embarrassingly,gowsmith2024word,batsuren2024evaluating}.

Research concerning inference methods has been more scarce, and includes examination of random effects on BPE merges~\cite{provilkov-etal-2020-bpe,saleva-lignos-2023-changes} and application of sophisticated search algorithms~\cite{he-etal-2020-dynamic}.
As far as we know, there exists no comprehensive study comparing inference methods across a variety of vocabularies and sizes using diverse metrics.

In this work, we conduct a controlled experiment isolating the effects of inference methods over four tokenizers, introducing an evaluation suite aggregating intrinsic benchmarks from various theoretical realms.\footnote{We release our code and data at \scriptsize{\url{https://github.com/MeLeLBGU/tokenizers\_intrinsic\_benchmark}}.}
We find that greedy inference methods work surprisingly well for all four vocabularies across morphological and information-theoretic metrics.
Furthermore, we demonstrate that SaGe yields state-of-the-art performance according to morphological metrics, and that inference methods that minimize token count perform strongest by cognitive metrics.

\section{Inference Methods}
\label{sec:methods}
Let $\mathcal{V}$ denote a vocabulary of subword tokens and $w$ denote a \emph{word} (or `pretoken'), the output of a pretokenizer.
We define
\begin{math}
s(\mathcal{V},w) := (t_1,...,t_k)
\end{math}
as a segmentation of $w$ into $k$ subword tokens such that
\begin{math}
\forall i, t_i \in \mathcal{V}
\end{math}
and that the concatenation of
\begin{math}
t_1,...,t_k  
\end{math}
results in $w$.
We use the term \emph{segmentation} to denote the application of an \emph{inference method} on a text given a \emph{token vocabulary}, as well as its result.

\begin{table*}
    \centering
    \small
    \begin{tabular}{llrll}
    \toprule
        
        Resource & Type & Size & Reference & License \\
        \midrule
         LADEC & Morphological &7,804 &\citet{Gagn2019LADECTL}& CC BY-NC 4.0 DEED \\ MorphoLex &Morphological & 12,029 &\citet{SnchezGutirrez2018MorphoLexAD} & CC BY-NC-SA 4.0 DEED\\
         MorphyNet &Morphological & 219,410 &\citet{batsuren-etal-2021-morphynet}& CC BY-SA 3.0 DEED \\
         DagoBert &Morphological & 279,443 &\citet{hofmann-etal-2020-dagobert} &Not specified---citation based\\
         UniMorph &Morphological & 143,454 &\citet{batsuren-etal-2022-unimorph}& CC BY 4.0 DEED\\
         UnBlend &Morphological & 312 &\citet{pinter-etal-2020-will}& GPL-3.0 \\
         CompoundPiece &Morphological & 22,896 &\citet{minixhofer-etal-2023-compoundpiece}& Not specified---citation based\\
         Cognitive data &Cognitive& 55,867 &\citet{beinborn-pinter-2023-analyzing} &MIT\\
         tokenization-scorer &Information Theory& --- &\citet{zouhar-etal-2023-tokenization} &Not specified---citation based\\
        \bottomrule
    \end{tabular}
    \caption{Size, Reference and License details of the resources in our benchmark.}
    \label{tab:resources}

\end{table*}

Current widely-employed tokenization schedules couple together the tokenizer vocabulary with the inference method.
However, we advocate for decoupling them, as they are independent processes.
Specifically, given a fixed token vocabulary produced from pre-training data, one could subsequently use any applicable inference method for the task at hand.
Thus, in our experiments, we use various intrinsic metrics to analyze the impact and performance of the several classes of inference methods:

\paragraph{Greedy inference methods} only consider and produce one token at each step.
We test three greedy approaches:
\textbf{Longest prefix}, which WordPiece uses by default~\citep{Wu2016GooglesNM}, selects the longest token in $\mathcal{V}$ that is a prefix of $w$, and then continues to iteratively segment the remaining text.
\textbf{Longest suffix} selects the longest token that is a suffix of $w$ and continues iteratively~\citep{jacobs2022lost,bauwens2023thesis}.
Since this strategy diverges from English Morphology, we consider it an intriguing baseline for assessing the impact of linguistic structure on the inference method.
\textbf{Longest token} selects the longest token that is contained in $w$, adds it to the generated segmentation, and then iteratively segments each remaining character sequence.
This was proposed by \citet{hofmann-etal-2022-embarrassingly} to approximate words by their $k$ longest tokens. 
They showed that it preserves morphological structure of words and leads to performance gains on some downstream tasks.

\paragraph{Merge rules-based inference methods}
begin with a word's character sequence and iteratively apply token-forming merge rules learnt by the tokenizer at the vocabulary creation phase, until none can be applied. 
This is BPE's default inference mode.\footnote{While ostensibly also compatible with WordPiece, we found no implementation of the model that provides an ordered list of its merges.}
In our experiments we test two variants for BPE:
The \textbf{deterministic} merge strategy recursively applies the first applicable BPE merge rule by its order in the trained merge list.
\textbf{Dropout}~\citep{provilkov-etal-2020-bpe} applies each valid merge rule with probability $p$, leading to a regularization effect where rare tokens surface more often and their embeddings can be better trained.
It has been shown to improve machine translation performance.

\paragraph{Likelihood-based inference methods}
use individual likelihood values assigned to tokens in order to find a segmentation for $w$ where the total likelihood is maximized~\citep{kudo-2018-subword,he-etal-2020-dynamic}.
\textbf{Default} uses likelihood values learned during vocabulary construction and considers the likelihood of a segmentation to be the product of individual likelihoods (from which UnigramLM gets its name).
\textbf{Least tokens} assigns a constant likelihood value to all tokens, effectively selecting a segmentation where the number of tokens is minimized.
While not suggested so far as a standalone inference method, this obecjtive is proposed for both vocabulary training and inference in the PathPiece algorithm~\cite{schmidt2024tokenization}.

\section{Intrinsic Benchmark}
\label{sec:bench}

\begin{table*}
    \centering
    \small
    \begin{tabular}{llccccc}
    \toprule
        Vocab & Inference & Morphological  & Cognitive  & Rényi  & Tokens & Decoding \\
         & method & alignment &  plausibility & efficiency & per word & diff\\

        \midrule
        \multirow{6}{*}{BPE}
        & \cellcolor{red!10} longest prefix & .8584 & .3266  & .4482 & 1.4273 & .0502 \\
        & \cellcolor{yellow!10} longest suffix & .6467 & .3170 & .4482 &  1.4286 & .0417 \\
        & \cellcolor{blue!10} longest token & .\textbf{8738} & .3302 & .4474 & 1.4261 & .0484\\
        & \cellcolor{green!10} least tokens & .7544 & .3321 & .4476 & \textbf{1.4237} & .0382\\
        & \cellcolor{purple!10} \emph{det. merges} & .6309 &  .\textbf{3355} & .4482 & 1.4308 & --- \\
        & \cellcolor{gray!10} dropout merge & .6081 &  .2925 & .\textbf{4537} & 1.5793 & .1313 \\

         \midrule
         \multirow{4}{*}{WordPiece} & \cellcolor{red!10} \emph{longest prefix} & .\textbf{8488} & .3307 & \textbf{.4507} & 1.4430 & --- \\
         & \cellcolor{yellow!10} longest suffix & .6288 & .3198 & .4502 &  1.4435 & .0656\\
         & \cellcolor{blue!10} longest token  & .8466 & \textbf{.3332} & .4500 & 1.4411 & .0216 \\
         & \cellcolor{green!10} least tokens & .7342 & .3306 & .4401 & \textbf{1.4319} & .0682\\

        \midrule
        \multirow{5}{*}{UnigramLM}
        & \cellcolor{red!10} longest prefix & .\textbf{9222} & .2858 & \textbf{.3400} & 1.7577 & .1187 \\
        & \cellcolor{yellow!10} longest suffix & .7520 & .2690 & .2897 & 1.7624 &  .0516\\
        & \cellcolor{blue!10} longest token & .8845 &  .2948 & .3040 & 1.7353 & .0406  \\
        & \cellcolor{green!10} least tokens & .8982 & .\textbf{2953} & .2969 & \textbf{1.7219} & .0328 \\
        & \cellcolor{cyan!10} \emph{likelihood} & .9149 & .2937 & .2919 & 1.7314 & --- \\

         \midrule
         
        \multirow{5}{*}{SaGe} & \cellcolor{red!10} \emph{longest prefix} & .\textbf{9606} & .2581 & .\textbf{3217} & 1.9445 & --- \\
        & \cellcolor{yellow!10} longest suffix & .7370 &.2471 & .2832 & 1.9615 & .1704\\
        & \cellcolor{blue!10} longest token & .9236 & .2671 & .3027 & 1.9236 & .0887 \\
        & \cellcolor{green!10} least tokens & .9125 & \textbf{.2674} & .2944 & \textbf{1.8895} &.1318\\
         & \cellcolor{cyan!10} likelihood$^\dagger$ &  .9515 & .2664 & .2937 & 1.9156 & .1168\\
         \bottomrule
    \end{tabular}
    \caption{Intrinsic Benchmark results on a vocab size of 40k. `Default' decoding algorithms (used in vocabulary construction) in \emph{italics}. Not all methods are applicable to all tokenizers. \emph{Decoding diff} presents the share of pretokens in the MiniPile test set that are differently tokenized using the method, compared with the default.
    We present correlation scores for performance over the various metric families in \autoref{app:correlations}.\\
    $^\dagger$For SaGe, likelihood is only based on unigram scores obtained before further vocabulary ablation.}
    \label{tab:intrinsic}

\end{table*}

Some analyses of tokenizers rely on training language models or translation models and evaluating their performance on downstream tasks.
Using this process to isolate effects of tokenization hyperparameters, such as inference method, is both time- and resource-consuming, as well as unstable due to the introduction of multiple sources of randomness throughout the LM/TM pre-training and fine-tuning phases.
Few measures have been introduced that are intrinsic to vocabularies and their direct application to corpora, and fewer still avoid conflating the measures with the objectives used in the vocabulary construction process itself.
As a result, the body of work focused on improving tokenization schemes is still relatively small.

We create and release a benchmark made to intrinsically evaluate subword tokenizers.
We collected word-level datasets and information measures which have been shown, or hypothesized, to correlate with the performance of language models on various downstream tasks.
Details on these resources are provided in \autoref{tab:resources}.
At present, the benchmark is focused on the English language, although corresponding datasets exist for others as well.

\paragraph{Morphological alignment}
It is commonly assumed that, for a given tokenizer, alignment of word segments to morphological gold-standard segmentations is a predictor of the ability of a language model that uses the given tokenizer to represent words, especially `complex' ones that are made up of several roots or contain multiple morphological affixes~\cite{schick2019rare,nayak-etal-2020-domain,hofmann-etal-2021-superbizarre,gow-smith-etal-2022-improving}.
We follow \citet{gow-smith-etal-2022-improving} and evaluate our tokenizers's alignment with morphological annotations found in LADEC~\citep{Gagn2019LADECTL}, MorphoLex~\citep{SnchezGutirrez2018MorphoLexAD}, MorphyNet~\citep{batsuren-etal-2021-morphynet}, and DagoBert~\citep{hofmann-etal-2020-dagobert}.
We augment these datasets with morpheme segmentation data~\citep{batsuren-etal-2022-unimorph}, novel blend structure detection data~\citep{pinter-etal-2020-will}, and compound separation data~\citep{minixhofer-etal-2023-compoundpiece}.
The number of words in each resource can be found in \autoref{tab:resources}.
We compare the segmentations generated by the tokenizers with each inference method to gold-standard morphological segmentations using the metric introduced by~\citet{Creutz2004MorphemeSG}, and report the macro-averaged F$_1$ score over the different resources.

\paragraph{Cognitive Plausibility} We use the benchmark and data from \citet{beinborn-pinter-2023-analyzing} to measure the correlation of a tokenizer's output with the response time and accuracy of human participants in a lexical decision task, predicated on the hypothesis that a good tokenizer struggles with character sequences that humans find difficult, and vice versa.
We report the average of the absolute value correlation scores across the four linguistic setups (word/nonword $\times$ accuracy/response time).

\paragraph{Tokens distribution statistics} 
We report the Rényi efficiency of different segmentations across a corpus~\cite{zouhar-etal-2023-tokenization}.
This measure penalizes token distributions dominated by either very high- and/or very low-frequency tokens, and was shown to correlate strongly with BLEU scores for machine translation systems trained on the respective tokenizers.
Recent work~\citep{cognetta-etal-2024-two-counterexamples} reveals a misalignment between Rényi efficiency and downstream performance in certain cases, reinforcing the necessity of an evaluation suite grounded in diverse domains and disciplines, as advocated in this work.
We also measure the average number of tokens per word over a corpus, as a proxy for compression quality~\citep{galle-2019-investigating}.
We omit the popular measure of character-length distribution of the tokens in the vocabulary, as it does not vary with segmentation strategy.

Lastly, we report the proportion of pretokens that are segmented different from the default across our reference corpus.

\section{Experiments}
\label{sec:exp}
We evaluate inference methods for the following tokenizer vocabularies: BPE, UnigramLM, WordPiece and SaGe.
We use the train split of the MiniPile~\citep{kaddour2023minipile} dataset to construct the tokenizer vocabularies.
We train vocabularies of sizes 32,768, 40,960, and 49,152, using the HuggingFace Tokenizers library, with identical pre-tokenization, representing the text at byte level.
UnigramLM and SaGe require an initial vocabulary for their top-down algorithms; for the former, we used the default implementation of one million top n-grams, while SaGe was initialized with a 262K-size UnigramLM vocabulary.
This initial vocabulary also provided us with token likelihood scores for inference, although a more exact implementation would also incorporate the contextual SaGe objective.

Token distribution statistics measurements and decoding diff rates were computed over the test split of the MiniPile dataset.
We measure the Rényi efficiency using the tokenization-scorer package\footnote{\scriptsize \url{https://github.com/zouharvi/tokenization-scorer}} with $\alpha=2.5$.
For each tokenizer, all experiments ran within several minutes on a personal laptop computer, highlighting the usefulness of our benchmark as an efficient tool for in-loop hyperparamter tuning.

We present the results on our benchmark for the 40K vocabularies in \autoref{tab:intrinsic}.
Results for other sizes are presented in \autoref{app:sizes}.
A breakdown of individual evaluation subsets is provided in \autoref{app:breakdown}.

\paragraph{Inference methods}
Within each tokenizer, we find that the default (`intended') strategy is often outperformed by others on some measures.
We observe a significant difference in morphological alignment when using merge rules-based inference methods.
Qualitative analysis showed the findings illustrated in \autoref{tab:examples}, where early merge rules such as `i-n', `a-m', or `o-n' cross morphological boundaries.
We notice a similar trend for likelihood-based inference, where frequently-used tokens possess very high likelihood values, sometimes exceeding those of the gold-standard segments.
We find that the \emph{least tokens} strategy fares well not only on the token count metric, which is mostly by-design, but also on cognitive measures, suggesting an effect of human preference to minimal word segmentation.
Finally, we observe that likelihood-based inference performs poorly in terms of Rényi efficieny, contrary to its stated purpose.
\emph{Dropout}, on the other hand, performs well on this measure, in line with its goal.
\emph{longest suffix} performs poorly across the board, possibly due to the suffixing nature of the English language, which has complementarily been shown to affect character-level sequential modeling~\citep{pinter-etal-2019-character}.
Notably, all our key observations are consistent across vocabulary sizes, as shown in \autoref{app:sizes}.

\paragraph{Inter-tokenizer results}
Our results align with \citet{bostrom-durrett-2020-byte}'s finding that BPE is inferior to UnigramLM on morphology alignment.
However, we show that some of this gap can be attributed not to the vocabulary but to the inference method.
In addition, we find that SaGe is most aligned to morphology by a substantial margin, indicating that its contextualized objective succeeds in retaining meaningful tokens in the vocabulary during ablation.
It is important to note that our evaluation is limited to English, a language with relatively low morphological complexity. Previous studies have identified significant tokenization challenges in non-English languages~\citep{mager-etal-2022-bpe}.
Therefore, any definitive conclusions regarding the effectiveness of tokenization methods should ideally encompass a diverse array of languages.
BPE and WordPiece, optimized for compression, unsurprisingly perform well above the likelihood-based vocabularies on the information measures.
However, we note that this carries over to the cognitive benchmark as well, supporting \citet{beinborn-pinter-2023-analyzing}'s findings.

Finally, we note that the two likelihood-based vocabularies follow the exact same within-vocab trends, and those for the two information-based vocabularies are also very close.
This highlights the consistency and robustness of our benchmark, although some results are relatively close to each other, which can be expected considering that some inference methods do not change much of the token sequences (see rightmost column of \autoref{tab:intrinsic}).

\section{Conclusion}
\label{sec:conc}
In this work, we curated an aggregated benchmark for intrinsic evaluation of subword tokenizers and used it to show the importance of selecting an inference method suited for a vocabulary given a task.
Given its computational efficiency, we hope the benchmark can be used in LM training efforts as a fruitful first step to improve tokenization schemes, or to select inference methods on-line.
Concretely, our findings suggest that greedy inference is a good choice, especially for morphologically-motivated tasks, even for tokenizers trained on other objectives.
Considering its ease of implementation and faster inference, this is an encouraging finding.

In the future, we plan to examine the correlation between our benchmark and various downstream tasks, as well as expand our experimentation to other languages and new algorithms.

\section*{Limitations}
Our paper contains evaluation of models in the English language.
This was done mostly in order to focus this short paper's contribution, and to be able to control for as many possibly-confounding variables such as training data.
Nevertheless, a more complete followup would have to include attempts to replicate our findings on other languages, aiming for a set as diverse as possible mostly in terms of typology and script.

Our evaluation is limited to intrinsic measures.
While this makes development of tokenizers easier, we acknowledge that the body of work correlating success on these measures with performance of downstream models on end-tasks is incomplete.

\section*{Ethical Considerations}
Details for human annotation for the cognitive benchmark are documented in the source benchmark's paper~\cite{beinborn-pinter-2023-analyzing}, from which we took the data as-is.

\section*{Acknowledgments}
We would like to thank Charlie Lovering, Varshini Reddy, and Haoran Zhang for comments on early drafts of this paper.
We thank the anonymous reviewers for their comments on our submission.
This research was supported in part by the Israel Science Foundation (grant No. 1166/23) and by a Google gift intended for work on \emph{Meaningful Subword Text Tokenization}.

\bibliography{anthology,custom}

\newpage

\appendix

\section{Results on Different Vocabulary Sizes}
\label{app:sizes}

\autoref{tab:other_sizes} presents benchmark results on 32K-sized and 49K-sized vocabularies.

\begin{table*}
    \centering
    \small
    \begin{tabular}{llccccc}
    \toprule
         Vocab & Inference & Morphological  & Cognitive  & Rényi  & Tokens & Decoding \\
         & method & alignment &  plausibility & efficiency & per word & diff\\
         
        \midrule

        \multirow{6}{*}{BPE-32K}
        & \cellcolor{red!10} longest prefix & .8727 & .3122 & .4600 & 1.4511 & .0581 \\
        & \cellcolor{yellow!10} longest suffix & .6496 & .3018 & .4602 & 1.4530 & .0469\\
        & \cellcolor{blue!10} longest token & .8883 & .3152 & .4592 & 1.4498 & .0558 \\
        & \cellcolor{green!10} least tokens & .7607 &  .3174 & .4595 & 1.4469 & .0426\\
        & \cellcolor{purple!10} \emph{det. merges} & .6409 &  .3201 & .4603 & 1.4551 & ---\\
        & \cellcolor{gray!10} dropout merge &  .6149  & .2795 & .4656 & 1.6041 & .1316\\
        
        \midrule

        \multirow{6}{*}{WordPiece-32K} & \cellcolor{red!10} \emph{longest prefix} & .7819 & .3185 & .4630 & 1.4689 & ---\\
         & \cellcolor{yellow!10} longest suffix & .5084 & .3089 & .4626 & 1.4698 & .0744\\
         & \cellcolor{blue!10} longest token & .7764 &  .3212 &  .4622 & 1.4667 & .0243\\
         & \cellcolor{green!10} least tokens & .7394 & .3185 & .4508 & 1.4565 & .0769\\
        
        \midrule

        \multirow{5}{*}{UnigramLM-32K}
        & \cellcolor{red!10} longest prefix & .9278 & .2855 & .3574 & 1.7803 & .1171\\
        & \cellcolor{yellow!10} longest suffix & .7610 & .2679 & .2961 & 1.7838 & .0516\\
        & \cellcolor{blue!10} longest token & .8926 & .2930 & .3103 & 1.7534 & .0395\\
        & \cellcolor{green!10} least tokens & .9077 & .2937 & .3028 & 1.7418 & .0303\\
        & \cellcolor{cyan!10} \emph{likelihood} & .9206 & .2931 & .2985 & 1.7501 & ---\\
        
         \midrule
         
        \multirow{5}{*}{SaGe-32K} & \cellcolor{red!10} \emph{longest prefix} & .9613 & .2610 & .3454 & 1.9502 & ---\\
        & \cellcolor{yellow!10} longest suffix & .7449 & .2473 & .2914 & 1.9736 & .1653\\
        & \cellcolor{blue!10} longest token & .9348 & .2685 & .3113 & 1.9319 & .0822 \\
        & \cellcolor{green!10} least tokens & .9212 & .2691 & .3035 & 1.9084 & .1247\\
         & \cellcolor{cyan!10} likelihood & .9579 &  .2679 & .3026 & 1.9246 & .1098\\
        
        \midrule
        \midrule

        \multirow{6}{*}{BPE-49K}
        & \cellcolor{red!10} longest prefix & .8440 & .3371 & .4391 & 1.4104 & .0444 \\
        & \cellcolor{yellow!10} longest suffix & .6438 & .3279 & .4390 & 1.4112 & .0379\\
        & \cellcolor{blue!10} longest token & .8637 & .3404 & .4384 & 1.4094 & .0430\\
        & \cellcolor{green!10} least tokens & .7464 & .3421 & .4385 & 1.4072 & .0351 \\
        & \cellcolor{purple!10} \emph{det. merges} & .6208 & .3461 & .4390 & 1.4137 & ---\\
        & \cellcolor{gray!10} dropout merge & .5967 & .2996 & .4446 & 1.5610 & .1310\\
        
        \midrule

        \multirow{4}{*}{WordPiece-49K} & \cellcolor{red!10} \emph{longest prefix} & .7600 & .3398 & .4413 & 1.4245 & --- \\
         & \cellcolor{yellow!10} longest suffix & .5133 & .3309 & .4407 & 1.4247 & .0589 \\
         & \cellcolor{blue!10} longest token & .7598 & .3421 & .4406 & 1.4228 & .0194\\
         & \cellcolor{green!10} least tokens & .7261 & .3401 & .4319 & 1.4145 & .0615\\
        
        \midrule

        \multirow{5}{*}{UnigramLM-49K}
        & \cellcolor{red!10} longest prefix & .9157 & .2818 & .3467 & 1.7432 & .1190\\
        & \cellcolor{yellow!10} longest suffix & .7449 & .2669 & .2849 & 1.7486 & .0516 \\
        & \cellcolor{blue!10} longest token & .8750 & .2915 & .2994 & 1.7245 & .0416\\
        & \cellcolor{green!10} least tokens & .8908 & .2926 & .2924 & 1.7098 & .0345\\
        & \cellcolor{cyan!10} \emph{likelihood} & .9095 & .2911 & .2871 & 1.7201 & ---\\
        
        \midrule

        \multirow{5}{*}{SaGe-49K} & \cellcolor{red!10} \emph{longest prefix} & .9606 & .2566 & .3361 & 1.9414 & ---\\
        & \cellcolor{yellow!10} longest suffix &  .7355 & .2466 & .2783 & 1.9562 & .1735\\
        & \cellcolor{blue!10} longest token & .9200   &.2662 & .2975 & 1.9192 & .0912\\
        & \cellcolor{green!10} least tokens & .9053 & .2662 & .2893 & 1.8947 & .1353\\
         & \cellcolor{cyan!10} likelihood & .9455 & .2651 & .2887 & 1.9111 & .1194\\
         
         \bottomrule
    \end{tabular}
    \caption{Aggregated results on 32K and 49K vocabularies.}
    \label{tab:other_sizes}

\end{table*}

\section{Detailed Results}
\label{app:breakdown}
\autoref{tab:morphological} breaks down the results (for 40K) on individual morphological datasets composing our benchmark. \autoref{tab:cognitive} Provides the same for individual cognitive measures.
\begin{table*}
    \centering
    \small
    \begin{tabular}{llccccccc}
    \toprule
        Vocab & Inference & Ladec & Morpho- & Morphy- & Dago- & Uni- & UnBlend & Compound- \\
        & & & Lex & Net & Bert & Morph & & Piece\\

        \midrule
        \multirow{6}{*}{BPE}
        & \cellcolor{red!10} longest prefix & .9210 & .8091 & .8511 & .8013 & .9956 & .7404 & .8904 \\
        & \cellcolor{yellow!10} longest suffix & .9497 & .6222 & .6524 & .7116 & .0316 & .6095 & .9502 \\
        & \cellcolor{blue!10} longest token & .9147 & .8125 & .8953 & .8618 & .9705 & .7711 & .8905 \\
        & \cellcolor{green!10} least tokens & .9775 & .7401 & .8303 & .8539 & .2573 & .6489 & .9731 \\
        & \cellcolor{purple!10} \emph{det. merges} & .8160 & .6781 & .6132 & .6195 & .3233 & .6097 & .7568 \\
        & \cellcolor{gray!10} dropout merge & .7666 & .6557 & .5871 & .5953 & .3128 & .6213 & .7178 \\

         \midrule
         \multirow{4}{*}{WordPiece} & \cellcolor{red!10} \emph{longest prefix} & .9333 & .7625 & .9114 & .8659 & .9963 & .5569 & .9153 \\
         & \cellcolor{yellow!10} longest suffix & .9447 & .6005 & .6289 & .6844 & .1059 & .4838 & .9535 \\
         & \cellcolor{blue!10} longest token & .9275 & .7568 & .9124 & .8765 & .9666 & .5749 & .9112 \\
         & \cellcolor{green!10} least tokens & .9706 & .7132 & .8253 & .8032 & .2670 & .5897 & .9704 \\

        \midrule
        \multirow{5}{*}{UnigramLM}
        & \cellcolor{red!10} longest prefix & .9551 & .8800 & .9291 & .9087 & .9973 & .8553 & .9299 \\
        & \cellcolor{yellow!10} longest suffix & .9248 & .6387 & .8206 & .8407 & .2777 & .8076 & .9536 \\
        & \cellcolor{blue!10} longest token & .8855 & .7534 & .9313 & .9378 & .9135 & .8571 & .9130 \\
        & \cellcolor{green!10} least tokens & .9660 & .8015 & .9511 & .9593 & .7218 & .9073 & .9801 \\
        & \cellcolor{cyan!10} \emph{likelihood} & .9341 & .7903 & .9645 & .9782 & .8423 & .9205 & .9743 \\

         \midrule
         
        \multirow{5}{*}{SaGe} & \cellcolor{red!10} \emph{longest prefix} & .9734 & .9422 & .9673 & .9600 & .9973 & .9213 & .9626 \\
        & \cellcolor{yellow!10} longest suffix & .9519 & .5996 & .7819 & .8091 & .2403 & .8216 & .9549 \\
        & \cellcolor{blue!10} longest token & 
.9420 & .8390 & .9365 & .9418 & .9711 & .8889 & .9457 \\
        & \cellcolor{green!10} least tokens & .9856 & .8394 & .9533 & .9632 & .7269 & .9318 & .9877 \\
         & \cellcolor{cyan!10} likelihood & .9709 & .8813 & .9809 & .9879 & .9014 & .9492 & .9890 \\
         \bottomrule
    \end{tabular}
    \caption{Results on individual morphological resources.}
    \label{tab:morphological}

\end{table*}
\begin{table*}
    \centering
    \small
    \begin{tabular}{llrrrr}
    \toprule
        Vocab & Inference & Words-RT & Words-ACC & nonwords-RT & nonwords-ACC\\
        \midrule

        \multirow{6}{*}{BPE}
        & \cellcolor{red!10} longest prefix &  $-$.3136 & .4035 & .4111 & $-$.1784 \\
        & \cellcolor{yellow!10} longest suffix & $-$.3102 & .3890 & .3987 & $-$.1699 \\
        & \cellcolor{blue!10} longest token & $-$.3164 & .4086 & .4130 & $-$.1828 \\
        & \cellcolor{green!10} least tokens & $-$.3146 & .4083 & .4226 & $-$.1828 \\
        & \cellcolor{purple!10} \emph{det. merges} & $-$.3285 & .4138 & .4163 & $-$.1835 \\
        & \cellcolor{gray!10} dropout merge & $-$.2562 & .3505 & .3908 & $-$.1726 \\

         \midrule
         \multirow{4}{*}{WordPiece} & \cellcolor{red!10} \emph{longest prefix} & $-$.3198 & .4029 & .4119 & $-$.1882  \\
         & \cellcolor{yellow!10} longest suffix & $-$.3132 & .3863 & .4028 & $-$.1770 \\
         & \cellcolor{blue!10} longest token & $-$.3226 & .4067 & .4134 & $-$.1902 \\
         & \cellcolor{green!10} least tokens & $-$.3146 & .4036 & .4201 & $-$.1842 \\

        \midrule
        \multirow{5}{*}{UnigramLM}
        & \cellcolor{red!10} longest prefix & $-$.2292 & .3391 & .3920 & $-$.1827 \\
        & \cellcolor{yellow!10} longest suffix & $-$.2308 & .3235 & .3645 & $-$.1572 \\
        & \cellcolor{blue!10} longest token & $-$.2493 & .3590 & .3904 & $-$.1804 \\
        & \cellcolor{green!10} least tokens & $-$.2394 & .3582 & .3978 & $-$.1860 \\
        & \cellcolor{cyan!10} \emph{likelihood} & $-$.2424 & .3577 & .3926 & $-$.1822  \\

         \midrule
         
        \multirow{5}{*}{SaGe} & \cellcolor{red!10} \emph{longest prefix} & $-$.1924 & .2896 & .3752 & $-$.1754  \\
        & \cellcolor{yellow!10} longest suffix & $-$.1895 & .2801 & .3602 & $-$.1585  \\
        & \cellcolor{blue!10} longest token & $-$.2079 & .3047 & .3790 & $-$.1767 \\
        & \cellcolor{green!10} least tokens & $-$.1978 & .3034 & .3864 & $-$.1821 \\
         & \cellcolor{cyan!10} likelihood & $-$.2035 & .3043 & .3797 & $-$.1780 \\
         \bottomrule
    \end{tabular}
    \caption{A breakdown of cognitive correlation results across vocabularies and inference methods.}
    \label{tab:cognitive}

\end{table*}

\section{Inter-Metric Correlations}
\label{app:correlations}
\autoref{tab:correlations} presents the Pearson correlation coefficients between the various intrinsic metrics used in the benchmark. These correlations are calculated based on the aggregated results across all vocabulary sizes.
\begin{table*}
    \centering
    \small
    \begin{tabular}{ccccc}
    \toprule
         & Morphological  & Cognitive  & Rényi  & Tokens\\
         & alignment &  plausibility & efficiency & per word \\
        \midrule
         Morphological alignment & 1 & $-$.5009 & $-$.4799 & .5726 \\
         Cognitive plausibility & --- & 1 & .6470 & $-$.9588 \\
         Rényi efficiency  & --- & --- & 1 & $-$.6400 \\
         Tokens per word   & --- & --- & --- & 1  \\
         \bottomrule
    \end{tabular}
    \caption{Correlations between the different intrinsic metrics.}

   \label{tab:correlations}

\end{table*}

\end{document}